\documentclass[runningheads]{llncs}
\usepackage[T1]{fontenc}
\usepackage{cite}
\usepackage{enumitem}
\setlist[itemize]{label=\textbullet}
\usepackage{color}

\newcommand{\bb}{\color{black}}
\usepackage{comment}
\usepackage{graphicx}
\usepackage{array, booktabs} 
\newcolumntype{C}[1]{>{\centering\arraybackslash}p{#1}}
\usepackage{calc}
\usepackage{adjustbox} 
\usepackage{multirow} 
\usepackage{makecell}
\usepackage{subcaption}
\usepackage{siunitx} 

\begin{document}
\renewcommand{\thefootnote}{\fnsymbol{footnote}}
\footnotetext[1]{\textsuperscript{} These authors contributed equally to this work.}
\title{Cross-dataset Multivariate Time-series Model for Parkinson’s Diagnosis via Keyboard Dynamics}
\titlerunning{Time-series Model for Parkinson’s Detection via Typing}
%
\author{Arianna Francesconi\inst{1}\orcidID{0009-0003-6648-575X} \and
Donato Cappetta\inst{2} \and
Fabio Rebecchi\inst{2} \and
Paolo Soda\inst{1,3}\orcidID{0000-0003-2621-072X} \and
Valerio Guarrasi\inst{1}\orcidID{0000-0002-1860-7447}\textsuperscript{*} \and
Rosa Sicilia\inst{1}\orcidID{0000-0002-2513-0827}\textsuperscript{*}}

\authorrunning{A. Francesconi et al.}
%
\institute{
Unit of Artificial Intelligence and Computer Systems, Department of Engineering, Università Campus Bio-Medico di Roma, Rome, Italy. \email{arianna.francesconi@unicampus.it} 
\and
Eustema S.p.A., Research and Development Centre, Naples, Italy. 
\and
Department of Radiation Sciences, Biomedical Engineering, Umeå University, Umeå, Sweden. 
}
\maketitle              

\begin{abstract}
Parkinson’s disease (PD) presents a growing global challenge, affecting over 10 million individuals, with prevalence expected to double by 2040. Early diagnosis remains difficult due to the late emergence of motor symptoms and limitations of traditional clinical assessments. In this study, we propose a novel pipeline that leverages keystroke dynamics as a non-invasive and scalable biomarker for remote PD screening and telemonitoring. Our methodology involves three main stages: (i)~preprocessing of data from four distinct datasets, extracting four temporal signals and addressing class imbalance through the comparison of three methods; (ii) pre-training eight state-of-the-art deep‐learning architectures on the two largest datasets, optimizing temporal windowing, stride, and other hyperparameters; (iii) fine-tuning on an intermediate-sized dataset and perform external validation on a fourth, independent cohort. Our results demonstrate that hybrid convolutional–recurrent and transformer‑based models achieve strong external validation performance, with AUC-ROC scores exceeding 90\% and F1-Score over 70\%. Notably, a temporal convolutional model attains an AUC-ROC of 91.14\% in external validation, outperforming existing methods that rely solely on internal validation. These findings underscore the potential of keystroke dynamics as a reliable digital biomarker for PD, offering a promising avenue for early detection and continuous monitoring.

\end{abstract}
\section{Introduction}
Parkinson’s disease (PD) is one of the major challenges for global healthcare systems, with a growing impact in terms of both prevalence and socioeconomic costs. PD affects over 10 million people worldwide, with projections indicating a doubling of this population by 2040~\cite{su2025projections}. The primary issue is late diagnosis, which limits the effectiveness of therapeutic interventions and exacerbates functional decline. Indeed, the hallmark motor symptoms (bradykinesia, rigidity, tremor) emerge only after a neural loss of at least 50\%~\cite{alfalahi2022diagnostic}, resulting in unsatisfactory diagnostic accuracy in the early stages. The search for accessible, non-invasive, and scalable biomarkers has thus become a top priority. 

In PD, subclinical motor signs can be detected early through abnormalities in motor sequencing and force stability; however, traditional neuropsychological tests and qualitative clinical rating scales present inherent limitations, including evaluator dependency, ceiling and floor effects, and low sensitivity to subtle fluctuations~\cite{alfalahi2022diagnostic}. In this context, keyboard dynamics (KD) have emerged as promising digital biomarkers capable of translating manual dexterity tests (e.g., finger tapping) into passively collected data via everyday devices such as smartphones and keyboards~\cite{giancardo2016computer,iakovakis2018touchscreen}.
The integration of artificial intelligence (AI), particularly deep learning (DL) models, with human-device interaction data represents a methodological breakthrough~\cite{roy2023imbalanced,furia2023exploring}. KD offers a rich space of kinetic features (e.g., key flight time and typing rhythm) that can be correlated with specific motor phenotypes such as arrhythmokinesia (irregular rhythm of movement) or movement heteroscedasticity (inconsistent speed across repetitions), leading to variability in keystroke timestamps.
DL architectures such as convolutional neural networks (CNNs) and transformers, already validated in physiological signal analysis, have the potential to extract latent patterns from typing time series, thereby overcoming the limitations of traditional models based on handcrafted features~\cite{dhir2020identifying}. However, their application to KD in PD remains limited and largely unexplored, particularly in cross-dataset validation scenarios that are critical for clinical deployment. 

This work aims to explore the effectiveness of advanced DL models for PD detection through the analysis of KD, moving beyond proof‑of‑concept to a robust cross‑dataset validation. We integrate data from multiple public datasets collected in different settings, including both \textit{free-text} (self-generated content, e.g., emails) and \textit{fixed-text} (standardized phrases) tasks, adopting a cross-dataset approach. Our study defines a replicable framework for patient stratification and real-time longitudinal telemonitoring, paving the way for accessible and personalized diagnostic tools. We advance the field of digital PD diagnosis through three main contributions:
\begin{itemize}
    \item Extensive analysis of state-of-the-art DL architectures for multivariate time-series modeling of KD signals;
    \item Comparison of three strategies for handling data imbalance, including unbalanced dataset (baseline), random undersampling, and a novel ensemble-based method called IMBALMED~\cite{francesconi2025class}, which creates multiple and complementary undersampled subsets to improve model diversity;
    \item Robust integration and validation on three public datasets using a pre-training followed by a fine-tuning strategy. Additionally, external validation is performed on a smaller, independent fourth dataset to demonstrate the generalizability of the proposed framework.
\end{itemize}

The rest of this manuscript is organized as follows. Section~\ref{sec:Related work} first introduces key concepts necessary to understand the application context of the proposed methodology. Then, it presents state-of-the-art studies on KD for PD detection. Section~\ref{sec:Methods} presents our methodology and experimental setup. Section~\ref{sec:Results} reports and discusses the obtained results. Finally, Section~\ref{sec:Conclusion} provides concluding remarks and future directions.

\section{Related Work}\label{sec:Related work}
The diagnosis of PD through the analysis of KD primarily focuses on four temporal signals: Hold Time (HT), the interval between pressing and releasing a key; Flight Time (FT), the time between releasing one key and pressing the next; Press-Press Time (PP), the duration between two consecutive key presses; and Release-Release Time (RR), the interval between two consecutive key releases.
KD data can be collected in two distinct contexts: \textit{in-the-clinic}, where data is acquired in controlled environments (e.g., laboratories or clinics), and \textit{in-the-wild}, where it is derived from daily interactions with personal devices (e.g., keyboards or smartphones) in uncontrolled settings.
In addition to being a highly imbalanced classification problem, since datasets typically contain fewer PD cases than controls, KD data introduce further challenges: they consist of multivariate, non-periodic time series, which complicates the application of standard balancing techniques. Time-domain oversampling or data augmentation methods often fail to preserve the temporal structure, whereas noise-based magnitude-domain augmentations risk altering meaningful motor patterns~\cite{iwana2021empirical}.

This section reviews studies that employ public KD datasets for PD diagnosis, categorizing them by learning approach (machine learning vs. deep learning) and highlighting if and how they address class imbalance.
To the best of our knowledge, the first study to employ KD for PD detection was conducted by Giancardo et al.~\cite{giancardo2016computer}. The authors introduced the public neuroQWERTY MIT-CSXPD dataset, consisting of 85 participants (42 with PD), whose data were collected \textit{in-the-clinic} via a touchscreen keyboard. Participants performed a \textit{free-text} task, typing spontaneously as they would at home. The study implemented an ensemble of 200 support vector regression models in a cross-validation setting, achieving an AUC-ROC of 79\%.
Subsequent research followed two main directions: traditional machine learning (ML) approaches and, more recently, DL methods. Among traditional ML approaches, Iakovakis et al~\cite{iakovakis2018touchscreen} used a public dataset (hereafter referred to as TyPD), which includes 33 participants (18 with PD), collected during a routine clinic visit using a smartphone-based \textit{fixed-text} task. The authors developed a two-stage ML model based on low- and high-order statistical features, such as mean, standard deviation, skewness, and kurtosis, extracted from HT, FT, and normalized pressure signals. Despite the limited sample size, their method achieved an AUC-ROC of 92\% using leave-one-patient-out cross-validation (LOPO).
Milne et al.~\cite{milne2018less}, using the neuroQWERTY MIT-CSXPD dataset, proposed a univariate logistic regression model based on mean absolute consecutive difference, a dynamic feature derived from the HT signal, which reached an AUC-ROC of 85\% under cross-validation.
More recently, Roy et al.~\cite{roy2023imbalanced} explicitly addressed class imbalance by applying undersampling to both the neuroQWERTY MIT-CSXPD and Online English datasets. The latter, composed of 230 participants (100 with PD), was collected \textit{in-the-wild} during a \textit{fixed-text} task using keyboards and involving multiple sentences. The authors combined ensembles of ML models (i.e., SVM) and DL models (i.e., long short-term memory (LSTM~\cite{hochreiter1997long})), extracting temporal features (i.e., mean, standard deviation, median, Q1, Q3) from HT, PP, and RR signals using sliding windows of 100 and 50 keystrokes for \textit{free-} and \textit{fixed-text} tasks, respectively. Their LOPO-based evaluation achieved an AUC-ROC of 85\%.

Regarding DL approaches, Iakovakis et al.~\cite{iakovakis2019early} employed the TyPD dataset in a study that implemented a 1D CNN with three convolutional layers, using input windows of 100 samples from HT and FT signals. Their model, evaluated with LOPO, achieved an AUC-ROC of 89\%.
Dhir et al.~\cite{dhir2020identifying} focused on HT and FT signals from the Online English dataset. They segmented with sliding windows of 50 characters, and applied an LSTM network validated via a hold-out strategy, reaching an AUC-ROC of 73\%. 
Bernardo et al.~\cite{bernardo2022modified} used the Tappy dataset (103 participants, 57 with PD) along with the Synthetic Minority Oversampling Technique (SMOTE) for balancing. The Tappy data were collected \textit{in-the-wild}, with participants using their personal computer keyboards to perform \textit{free-text} typing over longitudinal periods. They proposed a pipeline that transformed time-series data into spectrograms via continuous wavelet transform. To address the imbalance, they applied SMOTE for oversampling in the image domain. Their model, based on a SqueezeNet architecture and hold-out validation, reported an accuracy of 90\%, though the AUC-ROC was not specified.

The reviewed studies differ across datasets in terms of acquisition context (\textit{in-the-wild} vs. \textit{in-the-clinic}) and interaction device (keyboard vs. touchscreen). While such variability in training could foster generalization, most works rely on a single homogeneous dataset, limiting evaluation across heterogeneous sources or task settings. 
Although early results are encouraging, the field still faces several critical gaps, including the absence of approaches that are both effective and clinically scalable. 
First, few studies perform a comprehensive benchmarking of DL architectures specifically tailored to KD as time-series signals. 
Second, the problem of class imbalance is often overlooked or addressed through basic strategies. 
Third, even though some works explore typing on either keyboards or smartphones, very few attempt to integrate signals from both modalities, limiting the potential for widespread applicability. 
Lastly, most studies rely on single-dataset validation, with limited testing across heterogeneous data sources or task settings (e.g., \textit{free-text} vs. \textit{fixed-text}, \textit{in-the-wild} vs. \textit{in-the-clinic}).
Our work addresses these limitations through a unified framework that integrates multivariate typing signals, compares state-of-the-art DL architectures and balancing methods, and validates the approach across diverse public datasets.

\begin{figure}
\includegraphics[width=\textwidth]{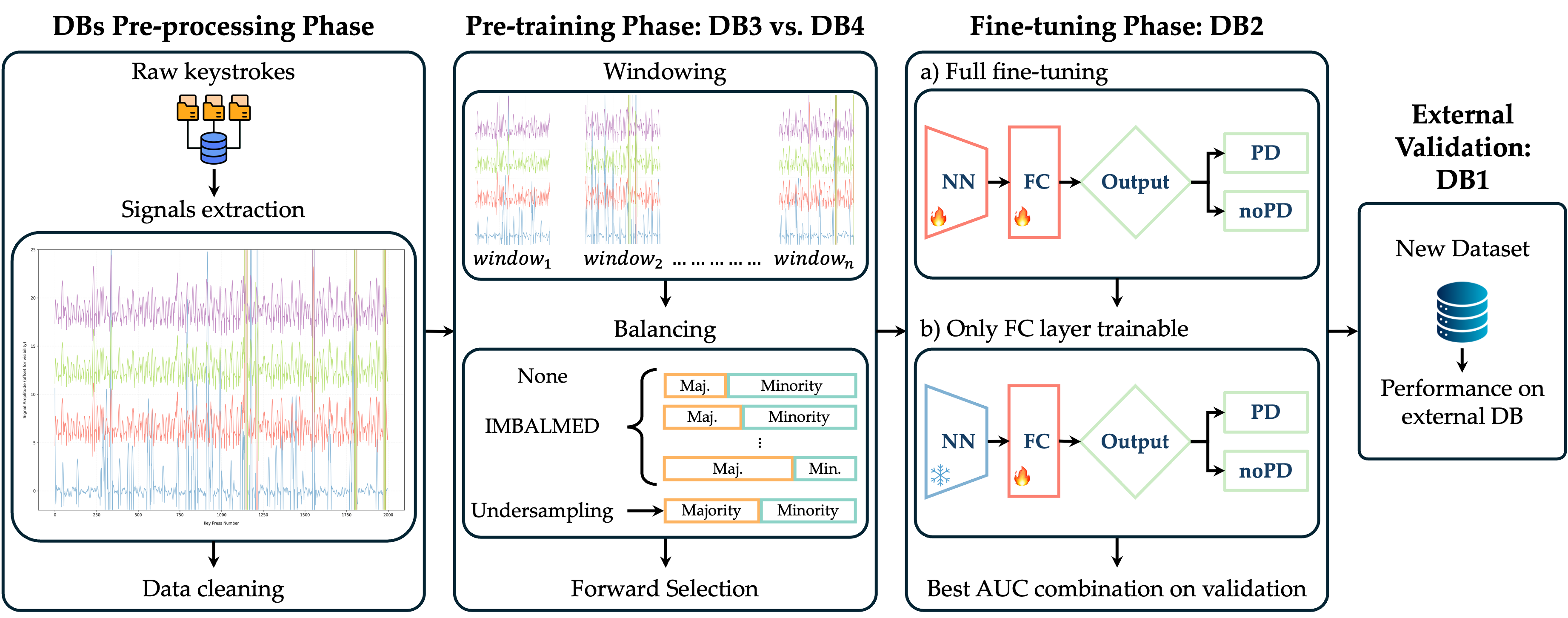}
\caption{Schematic representation of the proposed method, comprising four steps: dataset pre-processing, pre-training, fine-tuning, and external validation. In the NN (Neural Network) and FC (Fully Connected) blocks, weight freezing and updating are represented by ice and flame icons, respectively. Class balancing is illustrated with orange for the majority class (“Maj.”) and light blue for the minority class (“Min.”). In the IMBALMED~\cite{francesconi2025class} strategy, multiple sub-datasets with different class distributions are created.} \label{metodo}
\end{figure}

\section{Proposed Method and Experimental Setup}\label{sec:Methods}
The proposed methodology, illustrated in Figure~\ref{metodo}, consists of four main phases: pre-processing, pre-training, fine-tuning, and external validation. The following subsection details the pre-processing step; pre-training is presented in Section~\ref{subsec:pre-train}, whilst fine-tuning and external validation are described in Section~\ref{subsec:fine_ext}.

\subsection{Datasets and Pre-processing}\label{subsec:prepro}
We employed four publicly available datasets, described in detail in Section~\ref{sec:Related work}: TyPD (33 samples), neuroQWERTY MIT-CSXPD (85), Tappy (103), and Online English (230), which we refer to as DB1 through DB4, respectively, in order of increasing sample size. We focused on handcrafted temporal signals (HT, FT, PP, RR) instead of raw key timestamps, as these features are well-established in the KD–PD literature, as discussed in the previous section, and facilitate cross-dataset comparability.
Table~\ref{tab:datasets} provides an overview of these datasets, including: the dataset identifier~(ID) used in this work, the name and reference of the dataset (where the reference corresponds to the first paper that introduced and discussed the dataset), the number of participants with PD~(\#PD), the number of healthy controls~(\#HC), and the availability of four KD signals~(HT, FT, PP, and RR). Each of them is marked as Derived~(D) if computed from other signals or raw keystrokes, or Provided~(P) if directly available in the original dataset. 
The last four columns indicate the average signal length per subject~(in number of keystrokes) with the standard deviation, the average number of sessions per subject, the nature of the typing task~(\textit{Fixed-text} or \textit{Free-text}), and the context of data acquisition~(Clinic or Wild).
The average signal length was computed by aggregating the lengths of individual acquisition sessions for each subject within a dataset.
\renewcommand{\arraystretch}{1.5}  
\begin{table}
\centering
\caption{Summary of the four publicly available datasets used in this study, where `ID' stands for dataset identifier, `\#PD' is the number of PD  participants, and `\#HC' is the number of healthy controls. HT, FT, PP, and RR are the available KD signals marked as Derived~(D) or Provided~(P).
The last four columns report average signal length (measured in keystrokes per subject) (± SD), average sessions per subject, typing task (Fixed‑text vs. Free‑text), and acquisition context (Clinic vs. Wild).
}
\label{tab:datasets}
\begin{adjustbox}{width=\textwidth}
\begin{tabular}{c|c|c|c|c|c|c|c|c|c|c|c}
\toprule
\textbf{ID} & \textbf{Dataset Name} & \textbf{\#PD} & \textbf{\#HC} & \textbf{HT} & \textbf{FT} & \textbf{PP} & \textbf{RR} & \makecell{\textbf{Avg. Signal} \\ \textbf{Length}} & \makecell{\textbf{Avg. \#} \\ \textbf{Sessions}} & \textbf{Task} & \textbf{Context} \\
\hline
DB1 & TyPD~\cite{iakovakis2018touchscreen} & 18 & 15 & D & D & D & D & 60.18 ± 14.64 & 8.91 & \textit{Fixed-text} & Clinic \\
DB2 & \makecell{neuroQWERTY \\ MIT-CSXPD~\cite{giancardo2016computer}} & 42 & 43 & D & D & D & D & 1492.59 ± 644.76 & 1.36 & \textit{Free-text} & Clinic \\
DB3 & Tappy~\cite{goldberger2000physiobank} & 57 & 46 & P & P & P & D & 210.74 ± 82.34 & 145.09 & \textit{Free-text} & Wild \\
DB4 & Online English~\cite{dhir2020identifying} & 100 & 130 & D & D & D & D & 127.52 ± 22.15 & 14.57 & \textit{Fixed-text} & Wild \\
\bottomrule
\end{tabular}
\end{adjustbox}
\end{table}


The pre-processing steps were tailored to each dataset, following methodologies adopted in the already cited studies in which they were introduced or analyzed.
From each dataset, we extracted HT, FT, PP, and RR signals. For DB1 and DB4, both derived from \textit{fixed-text} tasks, we applied the preprocessing methodology proposed by~\cite{iakovakis2018touchscreen}: we enforced a minimum typing rate of 20 characters per minute per session to exclude recordings with irregular typing behavior. FT values exceeding 3 seconds were removed as outliers, whereas HT values were retained in their raw form, given their lower sensitivity to interruptions~\cite{iakovakis2018touchscreen}. The PP and RR signals were computed from the original press and release timestamps of each key. 
In contrast, DB2 and DB3 are based on \textit{free-text} tasks, and, in accordance with~\cite{bernardo2022modified}, no data cleaning was applied to preserve the natural typing rhythm of the users. However, a specific adjustment was necessary for DB3. This dataset, collected during the patients' daily activities, is organized on a monthly timescale, whereas all other datasets are structured daily. To harmonize the temporal resolution, we segmented the monthly data from DB3 into daily sessions. This segmentation was based not only on the date but also on the latency between consecutive keystrokes. When the time interval between two keystrokes exceeded 30 seconds, we considered this a discontinuity and treated it as the beginning of a new session. This threshold was chosen under the assumption that pauses longer than 30 seconds could lead to partial recovery from accumulated motor and cognitive fatigue, potentially disrupting the continuity of the typing pattern, which we aim to analyze through KD.
Finally, given the high average signal lengths across datasets (as reported in Table~\ref{tab:datasets}), we applied on-the-fly windowing during pre-training (see Section~\ref{subsec:pre-train}) to enable the model to capture both local patterns within each signal and global dependencies across the multivariate sequences.

The pre-processed datasets were then used in different phases of our experimental pipeline according to their size and task characteristics. Specifically, the two largest datasets, DB3 (\textit{free-text}) and DB4 (\textit{fixed-text}), were independently used as pre-training sources to assess how task type affects generalization performance. Their resulting models were then fine-tuned on the medium-sized DB2 (\textit{free-text}), allowing adaptation to a free-typing scenario. 
Finally, DB1, the smallest dataset based on a \textit{fixed-text} task, was used for external validation, providing an independent evaluation of the framework’s generalizability.
This experimental setting enabled us to assess not only the model's robustness across datasets of varying sizes, but also its ability to generalize across different typing task modalities, reflecting heterogeneous real-world usage conditions.
\bb

\subsection{Pre-training}\label{subsec:pre-train}
We adopted a pre-training strategy and evaluated eight DL architectures from the tsai library~\cite{tsai}, specifically designed for time series analysis. These models can be grouped into four main families: recurrent neural networks~(RNNs), including Gated Recurrent Unit~(GRU~\cite{cho2014learning}) and LSTM; hybrid models that combine recurrent and convolutional components~(RNN-CNN), namely GRU-Fully Convolutional Network~(GRU-FCN) and LSTM-Fully Convolutional Network~(LSTM-FCN)~\cite{karim2017lstm}; purely convolutional networks, such as Temporal Convolutional Network~(TCN~\cite{bai2018empirical}) and Explainable Convolutional Network~(XCM~\cite{fauvel2021xcm}); and transformer-based architectures, including Time-Series Transformer~(TSiT) and Time-Series Transformer Plus~(TSTPlus)~\cite{zerveas2021transformer}.

To investigate the effect of early stopping, we trained all models for 50 epochs and compared three patience settings: no patience, a patience of 5, and a patience of 25. We also compared two different loss functions: binary cross-entropy and focal loss. The latter was included to address class imbalance, as it reduces the influence of dominant classes by focusing on challenging minority examples.
Pre-training was conducted on the two largest datasets (DB3 and DB4), with each model trained independently. We did not merge DB3 and DB4 due to their task heterogeneity (\textit{free-text} vs. \textit{fixed-text}). Instead, we assessed them separately to analyze cross-context transferability. Hyperparameter optimization was conducted using a forward selection strategy: it was used as a simple grid-search-like strategy, where hyperparameters were optimized sequentially (first window size, then stride, etc.), fixing the best value from the previous step. For DB3, we explored window sizes of 90, 100, and 110 keystrokes; for DB4, we tested 40, 50, and 60, in accordance with~\cite{roy2023imbalanced}. For the stride parameter, we evaluated values of 1, half, and full window size. We tested batch sizes of 8, 16, and 32, and learning rates of 0.001, 0.0001, and 0.00001.
We further compared two checkpointing strategies: last epoch vs. best validation.
Finally, we examined the impact of different class balancing strategies: we compared no balancing, random undersampling, and the IMBALMED strategy~\cite{francesconi2025class}. IMBALMED (multImodal enseMble via class BALancing diversity for iMbalancEd Data) is an ensemble-based method that leverages undersampling to generate multiple imbalanced subsets, each characterized by a different class distribution. For instance, in the first subset shown in Figure~\ref{metodo}, the minority class represents only about 20\% of the data, but its proportion gradually increases across the subsets until it becomes the majority. This progressive diversification encourages the ensemble to learn from varying imbalance conditions, enhancing generalization. The method was previously shown to outperform other state-of-the-art balancing techniques on multimodal tabular data.
Its strong performance and compatibility with KD, where undersampling is particularly effective (see Section~\ref{sec:Related work}), motivated its inclusion in our experimental setup.

We used a leave-20\%-subjects-out cross-validation scheme (stratified by subject), which corresponds to a 10-fold partition but ensures subject-level separation, as some datasets contained multiple sessions per patient.

Performance was evaluated at the patient level by averaging predicted probabilities across windows and sessions and computing metrics on the aggregated outputs. 
The best-performing configuration for each model was selected based on validation performance for subsequent fine-tuning on a third dataset.

\subsection{Fine-tuning and External Validation}\label{subsec:fine_ext}
The second block of Figure~\ref{metodo} corresponds to the fine-tuning phase: the best-performing hyperparameter configuration was identified for each model and balancing strategy, based on their validation AUC-ROC. The selected configuration was subsequently used to fine-tune the classifiers on DB2, the dataset chosen for this stage due to its intermediate size.
A lower learning rate, set to an order of magnitude smaller than in pre-training, was used during fine-tuning to allow for more stable updates and to avoid disrupting the pre-trained representations.
We also compared two weight-freezing strategies: no frozen layers, allowing full fine-tuning, and freezing all layers except for the final fully connected layer. 
The best-performing weight-freezing strategy on the validation set was then used in the external validation phase.
All fine-tuning experiments were carried out using the same 10-fold cross-validation strategy presented in Section~\ref{subsec:pre-train}, aggregating the results across windows and sessions. 

For external validation, we assessed the models' configurations that achieved the highest validation performance during fine-tuning.
The entire DB1 dataset, the smallest one, was used as an independent test set to assess the generalizability of the proposed approach.  
This ensured that the most effective version of the model was tested on the independent DB1 dataset, providing a realistic estimate of out-of-distribution performance and supporting the assessment of model robustness in new clinical settings.

\section{Results and Discussions}\label{sec:Results}
For all pre-training experiments, selecting the best validation weights, applying an early-stopping patience of 5, and using focal loss consistently outperformed their respective alternatives. Therefore, only the results under these optimized settings are reported in this section.

Table~\ref{tab:hyperparams} summarizes the optimal hyperparameters selected on the validation set for each model across both DB3 and DB4, considering the three balancing strategies: Unbalanced, Undersample, and IMBALMED~\cite{francesconi2025class}. The table is organized in two parts: (a) reports the best window size (WS) and stride (ST); (b) shows the corresponding learning rate (LR) and batch size (BS) used during training. 
The optimal hyperparameters varied notably across model architectures and balancing strategies. IMBALMED generally enabled more flexible configurations, likely due to the regularizing effect introduced by the ensemble’s diversity. 
Recurrent models often favored shorter WS, while transformers sometimes benefited from longer WS, though results were not consistent across datasets (Table~\ref{tab:hyperparams}). 
Lower learning rates were typically selected under the undersampling strategy, likely reflecting the reduced data volume and the need for more cautious updates. 
Overall, these findings highlight that hyperparameter tuning must be carefully adapted to both the specific model and the balancing method employed.

\renewcommand{\arraystretch}{1.1}  
\begin{table}[t]
\centering
\caption{Optimal window size (WS), stride (ST), learning rate (LR), and batch size (BS) for each model, dataset, and balancing method, as determined during the pre-training step.}
\label{tab:hyperparams}
\begin{subtable}[t]{\textwidth}
\setlength{\tabcolsep}{8.8pt} 
\caption{Window size (WS) and stride (ST)}
\begin{adjustbox}{width=\textwidth}
\begin{tabular}{l|cc|cc|cc|cc|cc|cc}
\toprule
\multirow{3}{*}{\textbf{Model}} 
  & \multicolumn{6}{c|}{\textbf{DB3}} 
  & \multicolumn{6}{c}{\textbf{DB4}} \\
\cline{2-13}
  & \multicolumn{2}{c|}{\textbf{Unbal.}} 
  & \multicolumn{2}{c|}{\textbf{Under.}} 
  & \multicolumn{2}{c|}{\textbf{IMB.}} 
  & \multicolumn{2}{c|}{\textbf{Unbal.}} 
  & \multicolumn{2}{c|}{\textbf{Under.}} 
  & \multicolumn{2}{c}{\textbf{IMB.}} \\
\cline{2-13}
  & \textbf{WS} & \textbf{ST} 
  & \textbf{WS} & \textbf{ST} 
  & \textbf{WS} & \textbf{ST} 
  & \textbf{WS} & \textbf{ST} 
  & \textbf{WS} & \textbf{ST} 
  & \textbf{WS} & \textbf{ST} \\
\hline
GRU         & 90  & 45 & 90 & 90 & 90 & 90   & 50 & 1  & 50 & 1  & 40 & 40 \\
LSTM        & 100 & 50 & 90 & 45 & 90 & 45   & 40 & 20& 60 & 60 & 60 & 1  \\
GRU-FCN     & 90  & 45 & 100 & 100 & 90 & 90   & 60 & 60 & 40 & 40 & 40 & 1  \\
LSTM-FCN    & 90 & 90 & 90 & 45 & 90 & 90   & 40 & 1  & 50 & 1  & 50 & 1  \\
TCN         & 90 & 90 & 110 & 110 & 100 & 100   & 60 & 60 & 60 & 60 & 40 & 20\\
XCM         & 90 & 45 & 100 & 50 & 100 & 50   & 60 & 60 & 50 & 25 & 40 & 20\\
TSiT        & 90 & 90 & 90 & 90 & 100 & 100   & 40 & 40 & 40 & 20 & 60 & 60 \\
TSTPlus     & 90 & 90 & 90 & 45 & 100 & 100   & 60 & 60 & 60 & 60 & 50 & 50 \\
\bottomrule
\end{tabular}
\end{adjustbox}
\end{subtable}
\vspace{1em}
\begin{subtable}[t]{\textwidth}
\setlength{\tabcolsep}{5pt} 
\caption{Learning rate (LR) and batch size (BS)}
\begin{adjustbox}{width=\textwidth}
\begin{tabular}{l|cc|cc|cc|cc|cc|cc}
\toprule
\multirow{3}{*}{\textbf{Model}}
  & \multicolumn{6}{c|}{\textbf{DB3}} 
  & \multicolumn{6}{c}{\textbf{DB4}} \\
\cline{2-13}
  & \multicolumn{2}{c|}{\textbf{Unbal.}} 
  & \multicolumn{2}{c|}{\textbf{Under.}} 
  & \multicolumn{2}{c|}{\textbf{IMB.}} 
  & \multicolumn{2}{c|}{\textbf{Unbal.}} 
  & \multicolumn{2}{c|}{\textbf{Under.}} 
  & \multicolumn{2}{c}{\textbf{IMB.}} \\
\cline{2-13}
  & \textbf{LR} & \textbf{BS} 
  & \textbf{LR} & \textbf{BS} 
  & \textbf{LR} & \textbf{BS}
  & \textbf{LR} & \textbf{BS} 
  & \textbf{LR} & \textbf{BS} 
  & \textbf{LR} & \textbf{BS} \\
\hline
GRU         & 0.001 & 64 & 0.0001 & 8 & 0.001 & 32 & 0.001 & 16 & 0.0001 & 32 & 0.001 & 16 \\
LSTM        & 0.01 & 32 & 0.0001 & 8 & 0.001 & 8 & 0.001 & 16 & 0.001 & 16 & 0.001 & 16 \\
GRU-FCN     & 0.0001 & 32 & 0.001 & 16 & 0.001 & 32 & 0.001 & 16 & 0.001 & 16 & 0.001 & 16 \\
LSTM-FCN    & 0.001 & 32 & 0.0001 & 8 & 0.001 & 32 & 0.001 & 16 & 0.001 & 16 & 0.001 & 16 \\
TCN         & 0.0001 & 32 & 0.0001 & 8 & 0.001 & 64 & 0.001 & 16 & 0.001 & 64 & 0.001 & 16 \\
XCM         & 0.001 & 64 & 0.00001 & 8 & 0.001 & 32 & 0.0001 & 32 & 0.001 & 16 & 0.001 & 16 \\
TSiT        & 0.001 & 32 & 0.0001 & 16 & 0.01 & 32 & 0.001 & 16 & 0.001 & 16 & 0.001 & 16 \\
TSTPlus     & 0.001 & 32 & 0.0001 & 8 & 0.001 & 32 & 0.00001 & 64 & 0.001 & 32 & 0.001 & 16 \\
\bottomrule
\end{tabular}
\end{adjustbox}
\end{subtable}
\end{table}

\begin{table}[ht]
\centering
\caption{AUC-ROC scores comparison across DB3 and DB4 datasets for three balancing methods: Unbalanced (Unbal.), Undersample (Under.), and IMBALMED (IMB.)~\cite{francesconi2025class}. The highest score for each model is highlighted in bold.}
\begin{adjustbox}{width=\textwidth}
\begin{tabular}{@{}l|
                *{3}{C{1.5cm}}|
                *{3}{C{1.5cm}}@{}}
\toprule
\textbf{Model} & \multicolumn{3}{c|}{\textbf{DB3}} & \multicolumn{3}{c}{\textbf{DB4}} \\
\cline{2-7}
              & \textbf{Unbal.} & \textbf{Under.} & \textbf{IMB.} & \textbf{Unbal.} & \textbf{Under.} & \textbf{IMB.} \\
\hline
GRU        & 49.24\% & 65.93\% & 60.09\% & 64.51\% & 71.98\% & \textbf{73.27\%} \\
LSTM       & 46.06\% & 69.06\% & 58.49\% & 62.77\% & 59.75\% & \textbf{73.86\%} \\
GRU-FCN    & 58.79\% & 65.01\% & 63.80\% & 67.63\% & 67.73\% & \textbf{75.47\%} \\
LSTM-FCN   & 57.42\% & \textbf{71.28\%} & 62.60\% & 56.07\% & 64.71\% & 68.88\% \\
TCN        & 51.21\% & 57.82\% & 60.57\% & 63.42\% & 55.95\% & \textbf{70.49\%} \\
XCM        & 46.52\% & 64.33\% & 63.03\% & 58.41\% & 68.61\% & \textbf{71.55\%} \\
TSiT       & 48.12\% & \textbf{61.20\%} & 57.96\% & 50.00\% & 51.97\% & 55.51\% \\
TSTPlus    & 46.41\% & 63.90\% & 59.94\% & 57.94\% & 53.73\% & \textbf{67.32\%} \\
\bottomrule
\end{tabular}
\end{adjustbox}
\label{tab:pre_training}
\end{table}

Table~\ref{tab:pre_training} reports the AUC-ROC scores for each model on DB3 and DB4 under the three balancing strategies. For each model, the highest AUC-ROC value is highlighted in bold. As evident from the results, any balancing method boosts AUC-ROC over the unbalanced baseline. 
In the smaller, \textit{free-text} DB3, undersampling yields the largest gains (e.g., LSTM rises from 46\% to 69\%), likely by removing noisy majority samples. 
IMBALMED also improves performance but often trails undersampling on DB3 (e.g., GRU: 60\% vs 65.9\%), suggesting its diverse sub-dataset strategy could benefit from further tuning in this context. 
Conversely, on the larger \textit{fixed-text} DB4, IMBALMED almost always outperforms undersampling, indicating that creating multiple unbalanced subsets better captures intra-class variability when extensive data are available. 
Hybrid RNN-CNN models consistently lead pure RNNs, demonstrating that augmenting temporal recurrence with convolutional filters uncovers complementary kinetic patterns. 
Pure convolutional networks perform moderately: XCM approaches hybrid performance on DB4 (71.55\% vs top hybrids) but underperforms on DB3, showing they can still model temporal dependencies without recurrence. 
Attention-based models display the widest sensitivity to balancing: TSTPlus goes from 53.7\% unbalanced to 67.3\% under IMBALMED on DB4, implying they need further architectural or hyperparameter optimization to match more established RNN-CNN designs. 
Finally, absolute AUC-ROC values are higher on DB4 overall, reaffirming that larger \textit{fixed-text} corpora yield more stable keystroke biomarkers for PD detection.

\begin{table}
\centering
\caption{Fine-tuning performance comparison (AUC-ROC and F1-Score) on DB3 and DB4 using the IMBALMED~\cite{francesconi2025class} method. The highest AUC-ROC score for each model is highlighted in bold.}
\label{tab:fine_tuning}
\begin{adjustbox}{width=\textwidth}
\setlength{\tabcolsep}{12pt} 
\begin{tabular}{@{}l|
                *{2}{C{1.8cm}}|
                *{2}{C{1.8cm}}@{}}
\toprule
\textbf{Model} & \multicolumn{2}{c|}{\textbf{DB3 on DB2}} & \multicolumn{2}{c}{\textbf{DB4 on DB2}} \\
\cline{2-5}
              & \textbf{AUC-ROC} & \textbf{F1-Score}      & \textbf{AUC-ROC} & \textbf{F1-Score} \\
\hline
GRU           & 46.01\% & 63.87\% & \textbf{75.20\%} & 34.44\% \\
LSTM          & 55.87\% & 62.81\% & \textbf{81.50\%} & 54.56\% \\
GRU-FCN       & 37.71\% & 14.29\% & \textbf{84.52\%} & 88.87\% \\
LSTM-FCN      & 39.15\% & 36.11\% & \textbf{88.94\%} & 78.07\% \\
TCN           & 43.80\% & 51.02\% & \textbf{88.04\%} & 64.67\% \\
XCM           & 41.92\% & 41.98\% & \textbf{82.64\%} & 61.81\% \\
TSiT          & 66.61\% & 56.41\% & \textbf{85.29\%} & 64.44\% \\
TSTPlus       & 46.01\% & 18.33\% & \textbf{79.15\%} & 63.92\% \\
\bottomrule
\end{tabular}
\end{adjustbox}
\end{table}

In Table~\ref{tab:fine_tuning}, we report fine-tuning results for models pre-trained on DB3 and DB4, focusing exclusively on IMBALMED-balanced models, as this method led to higher AUC-ROC scores compared to undersampling for 6 out of 8 models (see Table~\ref{tab:pre_training}). 
Since full fine-tuning consistently yielded the best validation AUC-ROC across all experiments, we report only results obtained using this approach. 
Alongside AUC-ROC, we also include the F1-Score for a more complete evaluation. 
The highest AUC-ROC score for each model is highlighted in bold.
As the results show, transferring from DB3 to DB2 yields modest AUC-ROC values, peaking at around 66\%, whereas fine-tuning from the larger DB4 to DB2 produces substantially stronger performance, with top AUCs nearing 89\%. Although DB3 and DB2 are both free-text datasets, this counterintuitive gap may be explained by two factors: the higher noise inherent in DB3, collected in uncontrolled daily-life conditions with greater variability and distractions, and the larger size of DB4, which supports learning richer KD patterns and enhances transferability to DB2.
Within each transfer scenario, architectures that combine convolutional kernels with temporal recurrence (e.g., LSTM-FCN, GRU-FCN) consistently outperform pure RNNs or pure CNNs, suggesting that hybrid feature extractors best capture both local keystroke dynamics and longer-range dependencies. Finally, the optimal window and stride settings differ by source dataset: for DB3 (\textit{free-text}), shorter and non-overlapping windows performed best, likely because uncontrolled typing benefits from clean, independent segments. For DB4 (\textit{fixed-text}), mid-sized windows with small strides worked better, reflecting the structured and repetitive nature of the task.

\begin{table}
\centering
\caption{External validation results on DB1 after fine-tuning on DB2 and pre-training on DB4. In bold is the highest AUC-ROC for each model. The columns \textit{Best Fold} indicate the index of the fold from the fine-tuning cross-validation on DB2, whose checkpoint was used for external testing.}
\begin{adjustbox}{width=\textwidth}
\begin{tabular}{@{}l|
                *{3}{C{1.8cm}}|
                *{3}{C{1.8cm}}@{}}
\toprule
\textbf{Model} 
& \multicolumn{3}{c|}{\textbf{Pre-training on DB4}} 
& \multicolumn{3}{c}{\textbf{Fine-tuning on DB2}} \\
\textbf{} 
& \textbf{AUC-ROC} & \textbf{F1-Score} & \textbf{Best Fold} 
& \textbf{AUC-ROC} & \textbf{F1-Score} & \textbf{Best Fold} \\
\hline
GRU           & \textbf{88.15\%} & 45.45\% & 5 & 75.29\% & 56.13\% & 1 \\
LSTM          & \textbf{86.67\%} & 57.82\% & 1 & 84.31\% & 65.12\% & 9 \\
GRU-FCN       & 89.26\% & 78.26\% & 1 & \textbf{89.41\%} & 79.39\% & 1 \\
LSTM-FCN      & 82.96\% & 61.97\% & 2 & \textbf{90.32\%} & 80.42\% & 1 \\
TCN           & 81.85\% & 62.68\% & 2 & \textbf{91.14\%} & 79.39\% & 9 \\
XCM           & 70.37\% & 58.72\% & 10 & \textbf{73.33\%} & 66.67\% & 9 \\
TSiT          & 74.17\% & 57.91\% & 9 & \textbf{85.77\%} & 72.14\% & 1 \\
TSTPlus       & 62.75\% & 50.72\% & 1 & \textbf{90.59\%} & 73.91\% & 9 \\
\bottomrule
\end{tabular}
\end{adjustbox}
\label{tab:ext_val}
\end{table}

The results of the external validation on DB1 are reported in Table~\ref{tab:ext_val}, where we compare the performance of models using weights obtained after pre-training on DB4 (column \textit{Pre-training on DB4}) and after subsequent fine-tuning on DB2 (\textit{Fine-tuning on DB2}).  
We only report the results employing the pre-training and fine-tuning weights that yielded the best performance on the validation set during the 10-fold cross—validation. 
We show only  DB4 as the pre-training source, as it outperformed DB3 across all eight models when fine-tuned on DB2 (see Table~\ref{tab:fine_tuning}).
From the reported results, it can be seen that, even before fine-tuning, models pre-trained on DB4 already perform strongly on DB1; this is an expected outcome given that both datasets use the same \textit{fixed-text} protocol, which preserves similar keystroke patterns across domains. Fine-tuning on DB2 (a \textit{free-text} dataset) further improves performance by adapting these representations to more variable typing behavior without compromising the knowledge gained from \textit{fixed-text} pre-training. 
Among the architectures, TCN and TSTPlus emerged as the top performers, achieving AUC‑ROC scores of 91.14\% and 90.59\%, respectively, after fine‑tuning. Notably, TSTPlus, which initially underperformed compared to most models (see Table~\ref{tab:pre_training}), benefits substantially from fine‑tuning, underscoring the value of cross‑protocol adaptation. Hybrid recurrent-convolutional models also demonstrate strong performance by combining convolutional feature extraction with recurrent dynamics. In contrast, pure RNNs exhibit a performance drop on free‑text data, whilst TSiT and XCM register more modest gains. Overall, these findings confirm that leveraging a large \textit{fixed-text} corpus for pre-training, followed by targeted adaptation on \textit{free-text} data, yields the most generalizable keystroke-based biomarkers for PD detection. 

\section{Conclusion}\label{sec:Conclusion}
In this work, we have demonstrated the effectiveness of a four‐stage pipeline: pre‐processing, deep‐learning pre‐training, targeted fine‐tuning, and external validation for PD detection using KD. 
By leveraging large fixed‐text datasets~(DB4) for pre‐training and adapting the learned representations on a free‐text dataset (DB2), our framework achieves robust, generalizable performance.  
Hybrid recurrent and convolutional models~(e.g., GRU-FCN, LSTM-FCN) achieve strong results, confirming the value of combining local and temporal features~\cite{giancardo2016computer}. 
However, the TCN outperforms all, showing the effectiveness of deep temporal convolutions.

Our experiments further highlight the importance of tailored class‐imbalance handling: although simple undersampling yields strong gains on smaller free‐text corpora~(DB3), the IMBALMED~\cite{francesconi2025class} strategy enables superior generalization on the larger fixed‐text datasets by creating diverse, overlapping sub‐distributions that better represent intra‐class variability~\cite{ruffini2024multi, mogensen2025optimized}. Fine‐tuning on DB2 delivers a clear uplift over zero‐shot transfer from DB4, particularly rescuing transformer‐based models~(TSiT, TSTPlus) that initially underperform in the fixed‐text pre‐training phase.

External validation on the smallest independent dataset~(DB1) confirms the real-world applicability of our approach: LSTM-FCN, TCN, and TSTPlus models achieve AUC-ROC scores above 90\% and F1-Score exceeding 70\%. These results, superior to the current state of the art, which relies solely on external validation, underscore the potential of smartphone-based keystroke monitoring as a noninvasive, scalable biomarker for telemonitoring, supporting both remote PD screening and progression tracking.
Future work will explore the integration of multimodal sensor inputs~(e.g., accelerometry, voice)~\cite{di2025graph} and the incorporation of explainability techniques to enhance both diagnostic accuracy and model transparency~\cite{caragliano2025doctor, guarrasi2024multimodal}.

\begin{credits}
\subsubsection{\ackname} Arianna Francesconi is a Ph.D. student enrolled in the National Ph.D. in Artificial Intelligence, XXXIX cycle, course on Health and Life Sciences, organized by Università Campus Bio-Medico di Roma. This work was partially funded by: (i) PNRR – DM 117/2023; (ii) Eustema S.p.A.; (iii) PNRR MUR, Italy project PE0000013 - FAIR. Resources are provided by the National Academic Infrastructure for Supercomputing in Sweden (NAISS) and the Swedish National Infrastructure for Computing (SNIC) at Alvis @ C3SE.

\subsubsection{\discintname}
The authors have no competing interests to declare relevant to this article's content. 
\end{credits}

%
%
%
%

\end{document}